\pdfoutput=1

\documentclass[11pt]{article}

\usepackage[]{emnlp2021}

\usepackage{times}
\usepackage{latexsym}

\usepackage[T1]{fontenc}

\usepackage[utf8]{inputenc}

\usepackage{microtype}

\usepackage{times}
\usepackage{latexsym}
\usepackage[T1]{fontenc}
\usepackage[utf8]{inputenc}
\usepackage{microtype}
\usepackage{amsmath,amsthm,amssymb}
\usepackage{enumitem}
\usepackage{algorithmic}
\usepackage{mathtools}
\usepackage{multirow}
\usepackage{booktabs}
\usepackage{graphicx}
\usepackage{caption}
\usepackage{subcaption}
\usepackage[ruled, lined, linesnumbered, commentsnumbered, longend]{algorithm2e}

\SetCommentSty{mycommfont}

\DeclareMathOperator*{\argmax}{arg\,max}
\DeclareMathOperator*{\argmin}{arg\,min}

\newcommand{\method}{DIG}
\newcommand{\baselinea}{DIG-\textsc{RandomAnchor}}
\newcommand{\baselinen}{DIG-\textsc{RandomNeighbor}}
\newcommand{\maxcount}{\textsc{MaxCount}}
\newcommand{\greedy}{\textsc{Greedy}}

\DeclareMathAlphabet{\mathcal}{OMS}{cmsy}{m}{n}
\SetMathAlphabet{\mathcal}{bold}{OMS}{cmsy}{b}{n}

\newcommand{\bigo}{\mathcal{O}}

\title{Discretized Integrated Gradients for Explaining Language Models}

\author{Soumya Sanyal \and Xiang Ren \\
        University of Southern California \\
        \texttt{\{soumyasa, xiangren\}@usc.edu}}

\begin{document}
\maketitle
\begin{abstract}
As a prominent attribution-based explanation algorithm, Integrated Gradients (IG)
is widely adopted due to its desirable explanation axioms and the ease of gradient computation. It measures feature importance by averaging the model’s output gradient interpolated along a \textit{straight-line} path in the input data space. 
However, such straight-line interpolated points
are not representative of text data due to the inherent discreteness of the word embedding space. This questions the faithfulness of the gradients computed at the interpolated points and consequently, the quality of the generated explanations. Here we propose Discretized Integrated Gradients (DIG), which allows effective attribution along non-linear interpolation paths. 
We develop two interpolation strategies for the discrete word embedding space that generates interpolation points that lie close to actual words in the embedding space, yielding more faithful gradient computation.
We demonstrate the effectiveness of DIG over IG through experimental and human evaluations on multiple sentiment classification datasets. We provide the source code of \method{} to encourage reproducible research \footnote{\url{https://github.com/INK-USC/DIG}}.

\end{abstract}

\section{Introduction}
\label{sec:intro}

In the past few years, natural language processing has seen tremendous progress, largely due to strong performances yielded by pre-trained language models \cite{devlin2019bert,radford2019language,brown2020language}. But even with this impressive performance, it can still be difficult to understand the underlying reasoning for the preferred predictions leading to distrust among end-users \cite{lipton2018themythos}. 
Hence, improving model interpretability has become a central focus in the community with an increasing effort in developing methods that can explain model behaviors \cite{lime,binder2016layer,li2016understanding,sundararajan2017axiomatic,shrikumar2017learning,lundberg2017unified,murdoch2018beyond}.

\begin{figure}[t]
	\centering
	\includegraphics[width=0.85\columnwidth]{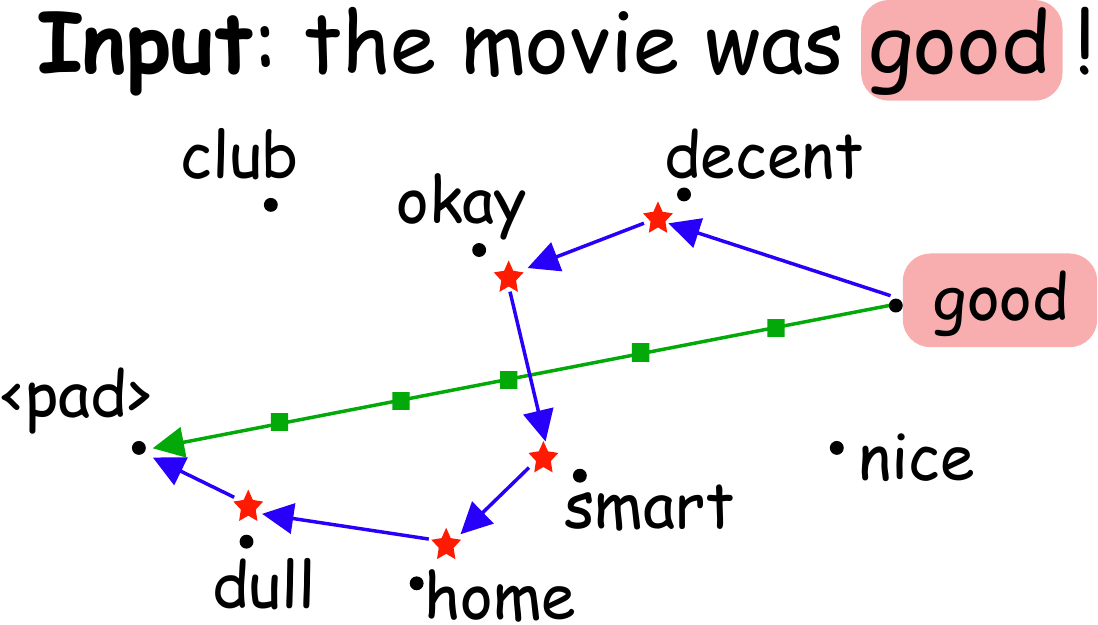}
	\caption{\textbf{An illustration of paths used in IG and \method{}}. IG uses a straight line interpolation with points as depicted by green squares. In contrast, \method{} uses a non-linear path (shown in blue) with interpolation points (red stars) lying close to words in the embedding space.}
	\label{fig:intro}
	\vspace{-0.2cm}
\end{figure}

Explanations in NLP are typically represented at a word-level or phrase-level by quantifying the contributions of the words or phrases to the model's prediction by a scalar score. These explanation methods are commonly referred as \textit{attribution-based methods} \cite{murdoch2018beyond,ancona2017towards}.
Integrated Gradients (IG) \cite{sundararajan2017axiomatic} is a prominent attribution-based explanation method used due to the many desirable explanation axioms and ease of gradient computation. It computes the partial derivatives of the model output with respect to each input feature as the features are interpolated along a straight-line path from the given input to a baseline value. For example, say we want to compute the attribution for the word ``\textit{good}'' in the sentence ``\textit{the movie was good!}'' using IG. The straight-line interpolation path used by IG is depicted in green in Figure \ref{fig:intro}. Here, the baseline word is defined as the ``\textit{<pad>}'' embedding and the green squares are the intermediate interpolation points in the embedding space.


While this method can be used for attributing inputs in both continuous (e.g., image, audio, etc.) and discrete (e.g., text, molecules, etc.) domains \cite{sundararajan2017axiomatic}, their usage in the discrete domain has some limitations. Since the interpolation is done along a straight-line path joining the input word embedding and the baseline embedding (``<pad>'' in Figure \ref{fig:intro}), the interpolated points are not necessarily representative of the discrete word embedding distribution. Specifically, let a dummy word embedding space be defined by the words represented by black dots in Figure \ref{fig:intro}. Then we can see that some of the green squares can be very far-off from any original word in the embedding space. Since the underlying language model is trained to effectively work with the specific word embedding space as input, using these out-of-distribution green interpolated samples as intermediate inputs to calculate gradients can lead to sub-optimal attributions.


\begin{figure*}
	\centering
	\begin{subfigure}{.5\textwidth}
		\centering
		\includegraphics[width=0.8\columnwidth]{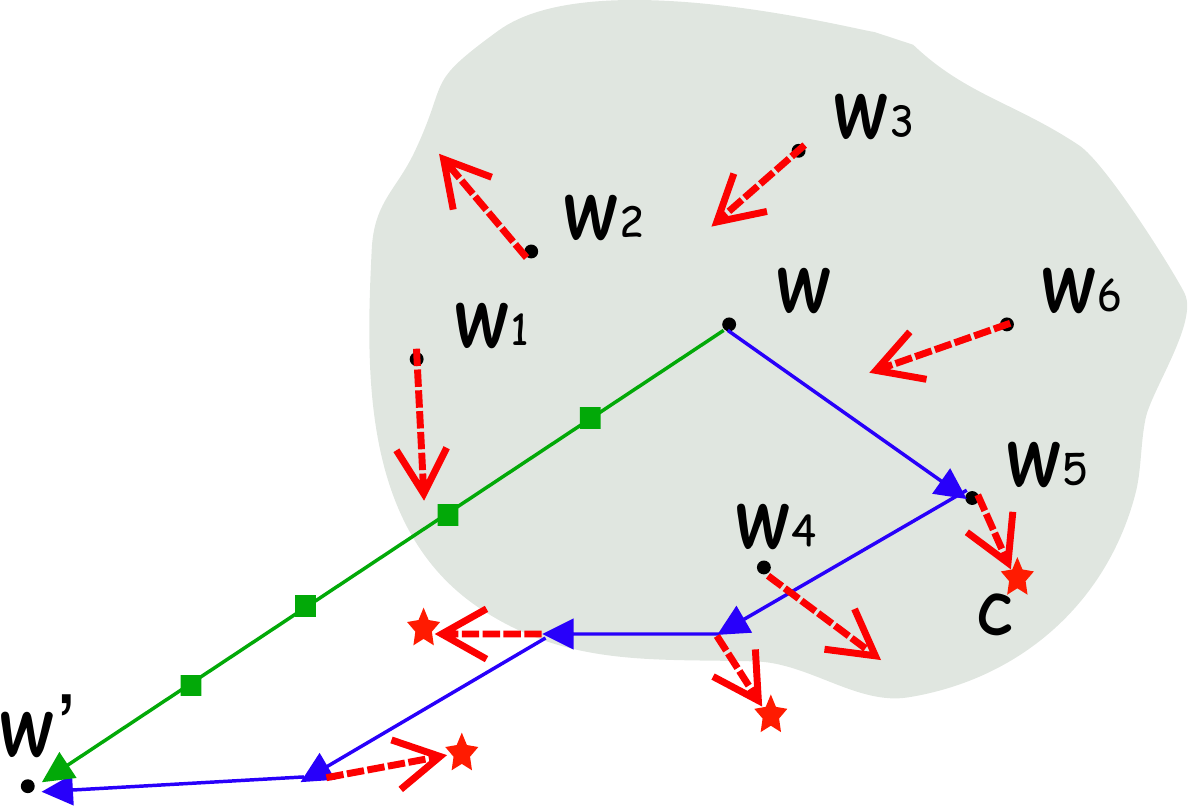}
		\caption{\method{}-\greedy{}}
		\label{fig:dig_greedy}
	\end{subfigure}%
	\begin{subfigure}{.5\textwidth}
		\centering
		\includegraphics[width=0.8\columnwidth]{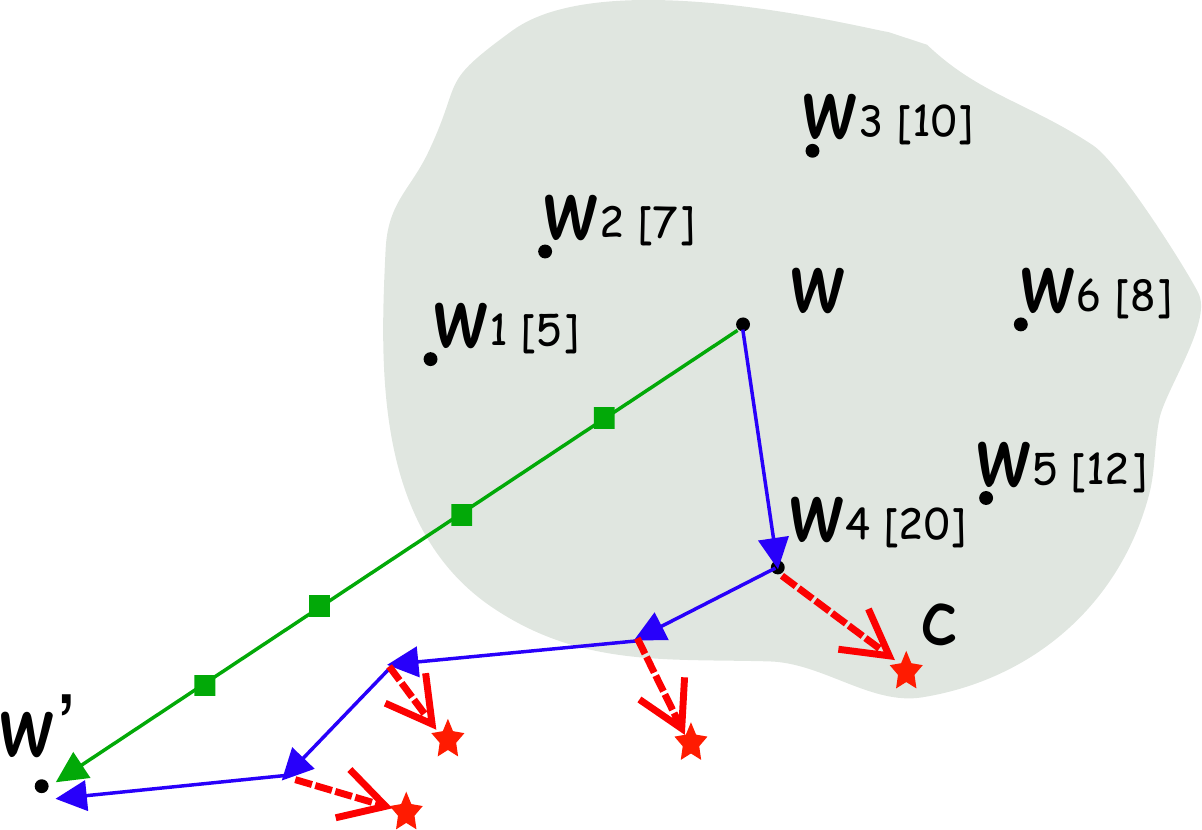}
		\caption{\method{}-\maxcount{}}
		\label{fig:dig_count}
	\end{subfigure}
	\caption{
	\textbf{Overview of paths used in \method{} and IG}. The gray region is the neighborhood of $w$. Green line depicts the straight-line path used by IG. \textbf{Left}: In \method{}-\greedy{}, we first monotonize each word in the neighborhood (red arrow) and the word closest to its corresponding monotonic point is selected as the anchor ($w_5$ since the red arrow of $w_5$ has the smallest magnitude). \textbf{Right}: In \method{}-\maxcount{} we select the word with the highest number of monotonic dimensions (count shown in $[.]$) as the anchor word ($w_4$), followed by changing the non-monotonic dimensions of $w_4$ (red arrow to $c$). Repeating this iteratively gives the non-linear blue path for \method{} with the red stars as interpolation points. Please refer to Section \ref{sec:dig} for more details. Figure best viewed in color.}
	\label{fig:dig}
\end{figure*}


To mitigate these limitations, we propose a Discretized integrated gradients (\method{}) formulation by relaxing the constraints of searching for interpolation points along a straight-line path. Relaxing this linear-path constraint leads to a new constraint on the interpolation paths in \method{} that points along the path should be monotonically situated between the input word embedding and the baseline embedding. Hence, in \method{}, our main objective is to monotonically interpolate between the input word embedding and baseline such that the intermediate points are close to real data samples. This would ensure that the interpolated points are more representative of the word embedding distribution, enabling more faithful model gradient computations. To this end, we propose two interpolation strategies that search for an optimal anchor word embedding in the real data space and then modify it such that it lies monotonically between the input word and baseline (see Fig.~\ref{fig:intro} for an illustration).

We apply \method{} using our proposed interpolation algorithms to generate attributions for three pre-trained language models - BERT \cite{devlin2019bert}, DistilBERT \cite{sanh2020distilbert}, and RoBERTa \cite{liu2019roberta}, each fine-tuned separately on three sentiment classification datasets - SST2 \cite{socher2013recursive}, IMDB \cite{mass2011learning}, and Rotten Tomatoes \cite{rotten_tomatoes}. We find that our proposed interpolation strategies achieve a superior performance compared to integrated gradients and other gradient-based baselines on eight out of the nine settings across different metrics. Further, we also observe that on average, end-users find explanations provided by \method{} to be more plausible justifications of model behavior than the explanations from other baselines.
\section{Method}
In this section, we first describe our proposed Discretized integrated gradients (\method{}) and the desirable explanation axioms satisfied by it. Then we describe an interpolation algorithm that leverages our \method{} in discrete textual domains.
Please refer to Appendix \ref{sec:background} for a brief introduction of the attribution-based explanation setup and the integrated gradients method.

\subsection{Discretized integrated gradients}
\label{sec:dig}
Below, we define our \method{} formulation that allows interpolations along non-linear paths:
\begin{equation}
\label{eq:dig_intergral}
\textrm{ \method{} }_{i}(x) = \int_{x^k_i=x^{\prime}_i}^{x_i} \frac{\partial F\left(x^k\right)}{\partial x_{i}} d x^k_i.
\end{equation}
Here, $x^k_i$ refers to the $i^{th}$ dimension of the $k^{th}$ interpolated point between input $x$ and  baseline $x^{\prime}$. The only constraint on $x^k_i$'s is that each interpolation should be monotonic between $x_i$ and $x^{\prime}_i$, i.e., $\forall j,k \in \{1, ..., m\}$; j < k,
\begin{equation}
\begin{split}
\label{eq:monotonic_constraints}
&x^{\prime}_i \leq x^j_i \leq x^k_i \leq x_i \textrm{~~if~~} x^{\prime}_i \leq x_i, \\
&x^{\prime}_i \geq x^j_i \geq x^k_i \geq x_i \textrm{~~otherwise}.
\end{split}
\end{equation}
This is essential because it allows approximating the integral in Eq. \ref{eq:dig_intergral} using Riemann summation\footnote{\url{https://en.wikipedia.org/wiki/Riemann\_sum}} which requires monotonic paths. We note that the interpolation points used by IG naturally satisfy this constraint since they lie along a straight line joining $x$ and $x^{\prime}$. The key distinction of our formulation from IG is that \method{} is agnostic of any fixed step size parameter $\alpha$ and thus allows non-linear interpolation paths in the embedding space.
The integral approximation of \method{} is defined as follows:
\begin{equation}
\label{eq:dig_sum}
\textrm{\method{}}^{\textrm{approx}}_{i}(x) = \Sigma_{k=1}^{m} \frac{\partial F\left(x^k\right)}{\partial x_{i}} \times \big(x^{k+1}_i - x^k_i\big),
\end{equation}
where $m$ is the total number of steps considered for the approximation.

\subsection{Axioms satisfied by \method{}}
\label{sec:axioms}
As described in prior works \cite{sundararajan2017axiomatic,shrikumar2017learning}, a good explanation algorithm should satisfy certain desirable axioms which justify the use of the algorithm for generating model explanations. Similar to IG, \method{} also satisfies many such desirable axioms. First, \method{} satisfies \textit{Implementation Invariance} which states that attributions should be identical for two \textit{functionally equivalent} models. Two models are functionally equivalent if they have the same output for the same input, irrespective of any differences in the model's internal implementation design. Further, \method{} satisfies  \textit{Completeness} which states that the sum of the attributions for an input should add up to the difference between the output of the model at the input and the baseline, i.e., $\sum_i \textrm{ \method{} }_{i}(x) = F(x) - F(x^{\prime})$. This ensures that if $F(x^{\prime}) \approx 0$ then the output is completely attributed to the inputs. Thirdly, \method{} satisfies \textit{Sensitivity} which states that attributions of inputs should be zero if the model does not depend (mathematically) on the input.
Please refer to Appendix \ref{sec:path} for further comparisons of \method{} with IG.

\subsection{Interpolation algorithm}
Here, we describe our proposed interpolation algorithm that searches for intermediate interpolation points between the input word embedding and the baseline embedding. Once we have the desired interpolation points, we can use Equation \ref{eq:dig_sum} to compute the word attributions similar to the IG algorithm. Please refer to Section \ref{sec:background_ig} for more details about application of IG to text.

\paragraph{Design Consideration.} First, we discuss the key design considerations we need to consider of our interpolation algorithm. Clearly, our interpolation points need to satisfy the monotonicity constraints defined in Equation \ref{eq:monotonic_constraints} so that we can use the Riemann sum approximation of \method{}. Hence, we need to ensure that every intermediate point lies in a monotonic path. Also, the interpolation points should lie close to the original words in the embedding space to ensure that the model gradients faithfully define the model behavior.
\\

Now, we define the notion of \textit{closeness} for our specific use-case of explaining textual models. To calculate how far the interpolated words are from some true word embedding in the vocabulary, we can compute the distance of the interpolated point from the nearest word in the vocabulary. We define this as the word-approximation error (WAE). More specifically, if $w^k$ denotes the $k^{th}$ interpolation point for a word $w$, then its word-approximation error along the interpolated path is defined as:
\begin{equation}
\label{eq:word_approx_error}
\textrm{WAE}_w = \frac{1}{m}\sum_{k=1}^{m} \min_{x \in V}dist(w^k - x),
\end{equation}
where $V$ is the embedding matrix of all the words in the vocabulary. WAE of a sentence is the average WAE of all words in the sentence. Intuitively, minimizing WAE will ensure that the interpolated points are close to some real word embedding in the vocabulary which in turn ensures that output gradients of $F$ are not computed for some out-of-distribution unseen embedding points.

We observe that to minimize WAE without the monotonic constraints defined in Section \ref{sec:dig}, one can define some heuristic to search for interpolation points that belong to the set $V$ (i.e., select words from the vocabulary as interpolation points), leading to a zero WAE. Motivated by this, for a given input word embedding, we first search for an anchor word from the vocabulary that can be considered as the next interpolation point. Since the anchor point need not be monotonic w.r.t. the given input, we then optimally perturb the dimensions of the anchor word so that they satisfy the monotonicity constraints in Equation \ref{eq:monotonic_constraints}. This perturbed point becomes our first interpolation. For subsequent interpolation points, we repeat the above steps using the previous anchor and perturbed points. Formally, we break our interpolation algorithm into two parts:
\begin{enumerate}[label=(\roman*)]
	\item \textsc{AnchorSearch}: In this step, given the initial word embedding $w$, we search for an anchor word embedding $a \in V$.

	\item \textsc{Monotonize}: This step takes the anchor embedding $a$ and modifies its dimensions to create a new embedding $c$ such that all dimensions of $c$ are monotonic between the input $w$ and the baseline $w^{\prime}$.
\end{enumerate}
Overall, given an initial input word embedding $w$ and a baseline embedding $w^{\prime}$, our interpolation algorithm interpolates points from $w$ to $w^{\prime}$ (which is in decreasing order of $k$ in Eq. \ref{eq:dig_sum}). It proceeds by calling \textsc{AnchorSearch} on $w$ to get an anchor word $a$. Then, it applies \textsc{Monotonize} on $a$ to get the monotonic embedding $c$. This is our first interpolated point (in reverse order), i.e., $c = w^{m-1}$. Now, the $a$ becomes the new $w$ for the next iteration and the process continues till $m$ steps. Next, we describe in detail our specific formulations of the \textsc{Monotonize} and \textsc{AnchorSearch} algorithms.
\\

\noindent\textsc{\textbf{Monotonize}}: In this step, given an anchor word embedding $a$, we modify the non-monotonic dimensions of $a$ such that they become monotonic w.r.t. $w$ and $w^{\prime}$. The monotonic dimensions of a vector $a$ is given by:
\begin{align*}
M_a &= \{j~|~ w^{\prime}_j \leq a_j \leq w_j, j \in \{1, ..., D\} \}\\
&\cup\{j ~|~ w^{\prime}_j \geq a_j \geq w_j, j \in \{1, ..., D\} \},
\end{align*}
where $D$ is the word embedding dimension. The number of monotonic dimensions is given by the size of the set defined as $|M_a|$. Thus, the non-monotonic dimensions $\overline{M_a}$ is the set complement of the monotonic dimensions, i.e., $\overline{M_a} = \{1, ..., D\} - M_a$, where the subtraction is the set-diff operation. Let the final monotonic vector be $c$. We define the \textsc{Monotonize} operations as follows:
\begin{align*}
c[M_a] &\leftarrow a[M_a],\\
c[\overline{M_a}] &\leftarrow w[\overline{M_a}] - \frac{1}{m} \times (w[\overline{M_a}] - w^{\prime}[\overline{M_a}]),
\end{align*}
where $m$ is the total number of interpolation points we want to select in the path. It can be easily seen that $c$ is monotonic w.r.t. $w$ and $w^{\prime}$ according to the definition in Equation \ref{eq:monotonic_constraints}.
\\

\noindent\textsc{\textbf{AnchorSearch}}: First, we preprocess the word embedding in $V$ to find the top-$K$ nearest neighbor for each word. We consider this neighborhood for candidate anchor selection. Let us denote the $K$-neighbors for a word $w$ by $KNN_V(w)$.
We define two heuristics to search for the next anchor word: \greedy{} and \maxcount{}.

In the \greedy{} heuristic, we first compute the monotonic embedding corresponding to each word in the neighborhood $KNN_V(w)$ using the \textsc{Monotonize} step. Then, we select the anchor word $a$ that is closest to its corresponding monotonic embedding obtained from the above step. This can be thought of as minimizing the WAE metric for a single interpolated word. The key intuition here is to locally optimize for smallest perturbations at each iterative selection step. This heuristic is depicted in Figure \ref{fig:dig_greedy} and the algorithm is presented in Algorithm \ref{algo:dig_greedy} in Appendix.

In the \maxcount{} heuristic, we select the anchor $a$ as the word in $KNN_V(w)$ with the highest number of monotonic dimensions. Precisely, the anchor is given by:
\begin{equation*}
a = \argmax_{a' \in KNN_V(w)} {|M_{a'}|}.
\end{equation*}
The intuition of this heuristic is that the vector with highest number of monotonic dimensions would require the minimum number of dimensions being perturbed in the \textsc{Monotonize} step and hence, would be close to a word in the vocabulary. This heuristic is depicted in Figure \ref{fig:dig_count} and the algorithm is presented in Algorithm \ref{algo:dig_count} in Appendix.
\begin{table*}[t]
	\centering
	\resizebox{\textwidth}{!}{%
		\begin{tabular}{lcccccccccccc}
			\toprule
			\multirow{2}{*}{\textbf{Method}}		& \multicolumn{4}{c}{\textbf{DistilBERT}}	& \multicolumn{4}{c}{\textbf{RoBERTa}} & \multicolumn{4}{c}{\textbf{BERT}}	\\
			\cmidrule(r){2-5} \cmidrule(r){6-9} \cmidrule(r){10-13}
			& LO $\downarrow$ & Comp $\uparrow$ & Suff $\downarrow$ & WAE $\downarrow$ & LO $\downarrow$ & Comp $\uparrow$ & Suff $\downarrow$ & WAE $\downarrow$ & LO $\downarrow$ & Comp $\uparrow$ & Suff $\downarrow$ & WAE $\downarrow$	\\
			\midrule
			Grad*Inp				& -0.402	& 0.112	& 0.375	& -		& -0.318	& 0.085	& 0.398	& -		& -0.502	& 0.168	& 0.366	& -	\\
			DeepLIFT				& -0.196	& 0.053	& 0.489	& -		& -0.300	& 0.078	& 0.432	& -		& -0.175	& 0.063	& 0.470	& -	\\
			GradShap				& -0.778	& 0.216	& 0.308	& -		& -0.523	& 0.168	& 0.347	& -		& -0.686	& 0.225	& 0.333	& -	\\
			IG 						& -0.950	& 0.248	& 0.275	& 0.344	& -0.738	& 0.222	& 0.250	& 0.669	& -0.670	& 0.237	& 0.396	& 0.302		\\
			\midrule
			\method{}-\greedy{} 	& -1.222	& \textbf{0.310}	& \textbf{0.237}	& 0.229	& -0.756	& 0.218	& \textbf{0.215}	& 0.460	& \textbf{-0.879}	& \textbf{0.292}	& \textbf{0.374}	& 0.249		\\
			\method{}-\maxcount{} 	& \textbf{-1.259}	& 0.307	& 0.241	& \textbf{0.227}	& \textbf{-0.826}	& \textbf{0.227}	& 0.238	& \textbf{0.439}	& -0.777	& 0.272	& 0.377	& \textbf{0.173}		\\
			\bottomrule
		\end{tabular}%
	}
	\caption{\label{tab:results_sst} Comparison of variants of \method{} with baselines on three LMs fine-tuned on SST2 dataset. `-' denotes that the WAE metric is not computable for that setting. We observe that \method{} outperforms the baselines on all three LMs. Please refer to Section \ref{sec:results} for more details.}
\end{table*}

\begin{table*}[t]
	\centering
	\resizebox{\textwidth}{!}{%
		\begin{tabular}{lcccccccccccc}
			\toprule
			\multirow{2}{*}{\textbf{Method}}		& \multicolumn{4}{c}{\textbf{DistilBERT}}	& \multicolumn{4}{c}{\textbf{RoBERTa}} & \multicolumn{4}{c}{\textbf{BERT}}	\\
			\cmidrule(r){2-5} \cmidrule(r){6-9} \cmidrule(r){10-13}
			& LO $\downarrow$ & Comp $\uparrow$ & Suff $\downarrow$ & WAE $\downarrow$ & LO $\downarrow$ & Comp $\uparrow$ & Suff $\downarrow$ & WAE $\downarrow$ & LO $\downarrow$ & Comp $\uparrow$ & Suff $\downarrow$ & WAE $\downarrow$	\\
			\midrule
			Grad*Inp				& -0.197	& 0.081	& 0.212	& -	& -0.195	& 0.043	& 0.279	& -	& -0.731	& 0.102	& \textbf{0.231}	& - 	\\
			DeepLIFT				& -0.021	& -0.009	& 0.534	& -	& -0.157	& 0.028	& 0.340	& -	& -0.335	& 0.023	& 0.486	& - 	\\
			GradShap				& -0.473	& 0.185	& 0.154	& -	& -0.416	& 0.129	& 0.196	& -	& -0.853	& 0.190	& 0.255	& - 	\\
			IG 						& -0.446	& 0.182	& 0.224	& 0.379	& \textbf{-0.733}	& \textbf{0.226}	& \textbf{0.084}	& 0.708	& -0.641	& 0.107	& 0.295	& 0.333		\\
			\midrule
			\method{}-\greedy{} 	& \textbf{-0.878}	& \textbf{0.319}	& \textbf{0.133}	& 0.256	& -0.683	& 0.198	& 0.100	& \textbf{0.484}	& \textbf{-1.152}	& \textbf{0.221}	& 0.240	& 0.287		\\
			\method{}-\maxcount{} 	& -0.795	& 0.296	& 0.152	& \textbf{0.255}	& -0.470	& 0.121	& 0.213	& 0.470	& -0.995	& 0.195	& 0.245	& \textbf{0.190}		\\
			\bottomrule
		\end{tabular}%
	}
	\caption{\label{tab:results_imdb} Comparison of variants of \method{} with baselines on three LMs fine-tuned on IMDB dataset. We observe that \method{} outperforms the baselines on DistilBERT and BERT models. Please refer to Section \ref{sec:results} for more details.}
\end{table*}

\section{Experimental Setup}
In this section, we describe the datasets and models used for evaluating our proposed algorithm.

\paragraph{Datasets.} The SST2 \cite{socher2013recursive} dataset has 6920/872/1821 example sentences in the train/dev/test sets. The task is binary classification into positive/negative sentiment. The IMDB \cite{mass2011learning} dataset has 25000/25000 example reviews in the train/test sets with similar binary labels for positive and negative sentiment. Similarly, the Rotten Tomatoes (RT) \cite{rotten_tomatoes} dataset has 5331 positive and 5331 negative review sentences. We use the processed dataset made available by HuggingFace Dataset library \footnote{\url{https://github.com/huggingface/datasets}} \cite{wolf-2020HuggingFace-datasets}.


\paragraph{Language Models.} We use pre-trained BERT \cite{devlin2019bert}, DistilBERT \cite{sanh2020distilbert}, and RoBERTa \cite{liu2019roberta} text classification models individually fine-tuned for SST2, IMDB, and RT datasets. The fine-tuned checkpoints used are provided by the HuggingFace library \cite{wolf-etal-2020-transformers}.

\paragraph{Evaluation Metrics.}
Following prior literature, we use the following three automated metrics:
\begin{itemize}
	\item \textbf{Log-odds} (LO) score \cite{shrikumar2017learning} is defined as the average difference of the negative logarithmic probabilities on the predicted class before and after masking the top $k\%$ words with zero padding. Lower scores are better.
	\vspace{-0.1cm}
	
	\item \textbf{Comprehensiveness} (Comp) score \cite{eraser} is the average difference of the change in predicted class probability before and after removing the top $k\%$ words. Similar to Log-odds, this measures the influence of the top-attributed words on the model's prediction. Higher scores are better.
	\vspace{-0.1cm}

	\item \textbf{Sufficiency} (Suff) score \cite{eraser} is defined as the average difference of the change in predicted class probability before and after keeping only the top $k\%$ words. This measures the adequacy of the top $k\%$ attributions for model's prediction.
	\vspace{-0.1cm}
\end{itemize}


Please refer to Appendix \ref{sec:eval_metrics_details} for more details about the evaluation metrics. We use $k=20\%$ in our experiments. In Appendix \ref{sec:topk} we further analyze the effect of changing top-k\% on the metrics. Additionally, we use our proposed word-approximation error (WAE) metric to compare \method{} with IG.

\begin{table*}[]
	\centering
	\resizebox{\textwidth}{!}{%
		\begin{tabular}{lcccccccccccc}
			\toprule
			\multirow{2}{*}{\textbf{Method}}		& \multicolumn{4}{c}{\textbf{DistilBERT}}	& \multicolumn{4}{c}{\textbf{RoBERTa}} & \multicolumn{4}{c}{\textbf{BERT}}	\\
			\cmidrule(r){2-5} \cmidrule(r){6-9} \cmidrule(r){10-13}
			& LO $\downarrow$ & Comp $\uparrow$ & Suff $\downarrow$ & WAE $\downarrow$ & LO $\downarrow$ & Comp $\uparrow$ & Suff $\downarrow$ & WAE $\downarrow$ & LO $\downarrow$ & Comp $\uparrow$ & Suff $\downarrow$ & WAE $\downarrow$	\\
			\midrule
			Grad*Inp				& -0.152	& 0.068	& 0.315	& -	& -0.158	& 0.054	& 0.406	& -	& -0.801	& 0.204	& 0.398	& - \\
			DeepLIFT				& -0.077	& 0.017	& 0.372	& -	& -0.150	& 0.050	& 0.413	& -	& -0.388	& 0.096	& 0.438	& - \\
			GradShap				& -0.298	& 0.156	& 0.270	& -	& -0.290	& 0.128	& 0.338	& -	& -0.809	& 0.235	& 0.388	& - \\
			IG 						& -0.424	& 0.208	& 0.189	& 0.348	& -0.368	& \textbf{0.149}	& 0.317	& 0.677	& -0.789	& 0.203	& 0.418	& 0.305		\\
			\midrule
			\method{}-\greedy{} 	& \textbf{-0.501}	& \textbf{0.257}	& \textbf{0.184}	& 0.329	& \textbf{-0.393}	& 0.148	& \textbf{0.294}	& 0.465	& \textbf{-1.056}	& \textbf{0.267}	& \textbf{0.416}	& 0.251		\\
			\method{}-\maxcount{} 	& -0.467	& 0.231	& 0.190	& \textbf{0.230}	& -0.361	& 0.133	& 0.332	& \textbf{0.444}	& -0.874	& 0.237	& 0.430	& \textbf{0.178}		\\
			\bottomrule
		\end{tabular}%
	}
	\caption{\label{tab:results_rotten} Comparison of variants of \method{} with baselines on three LMs fine-tuned on Rotten Tomatoes dataset. We observe that \method{} outperforms the baselines on all three LMs. Please refer to Section \ref{sec:results} for more details.}
\end{table*}

\section{Results}

\subsection{Performance Comparison}
\label{sec:results}
\vspace{-0.1cm}

We compare \method{} with four representative gradient-based explanation methods - Gradient*Input (Grad*Inp) \cite{shrikumar2016not}, DeepLIFT \cite{shrikumar2017learning}, GradShap \cite{lundberg2017unified}, and integrated gradients \cite{sundararajan2017axiomatic}. For the IMDB and RT datasets, we randomly sample a subset of 2,000 reviews from the public test sets to compare the different methods, due to computation costs. For the SST2 dataset, we use the complete set of 1,821 test sentences. The results are shown in Tables \ref{tab:results_sst}, \ref{tab:results_imdb}, and \ref{tab:results_rotten} for SST2, IMDB, and Rotten Tomatoes respectively.

\paragraph{Comparison with baselines.} First, we observe that across the nine different settings we studied (three language models per dataset), \method{} consistently outperforms the baselines on eight of the settings. This is valid for all the metrics. We also note that the WAE metric is lower for all variants of \method{} compared to IG. This validates that our proposed interpolation strategies for \method{} is able to considerably reduce the word-approximation error in the interpolated paths and consistently improving performance on all three explanation evaluation metrics considered.

\paragraph{Comparison between variants of \method{}.} Second, we observe that on average, \method{}-\greedy{} performs better than \method{}-\maxcount{}. Specifically, we find that \method{}-\maxcount{} doesn't outperform \method{}-\greedy{} by significantly large margins on any setting (while the opposite is true for one setting - RoBERTa fine-tuned on IMDB dataset). This could be because the \method{}-\greedy{} strategy ensures that the monotonic point $c$ is always close to the anchor $a$ due to the locally greedy selection at each step which is not explicitly guaranteed by \method{}-\maxcount{}. But overall, we do not find any specific performance trend between the two proposed variants and plan to study the influence of the embedding distribution in future works.

\paragraph{Analysis.} Finally, though we are able to achieve good reductions in WAE, we note that the WAE for our interpolation algorithms are not close to zero yet. This leaves some scope to design better interpolation algorithms in future. Moreover, we find that the average Pearson correlation between log-odds and WAE is 0.32 and the correlation is 0.45 if we consider the eight settings where we outperform IG. We discuss the correlations of all the settings in Appendix \ref{sec:correlation}. While this suggests a weak correlation between the two metrics, it is hard to comment if there is a causality between the two. This is partially because we believe selection of interpolation points should also take the semantics of the perturbed sentences into consideration, which we don't strongly enforce in our strategies. Hence, we think that constraining interpolations in a semantically meaningful way is a promising direction to explore.

\begin{figure}[t]
\vspace{-0.2cm}
	\centering
	\includegraphics[width=\columnwidth]{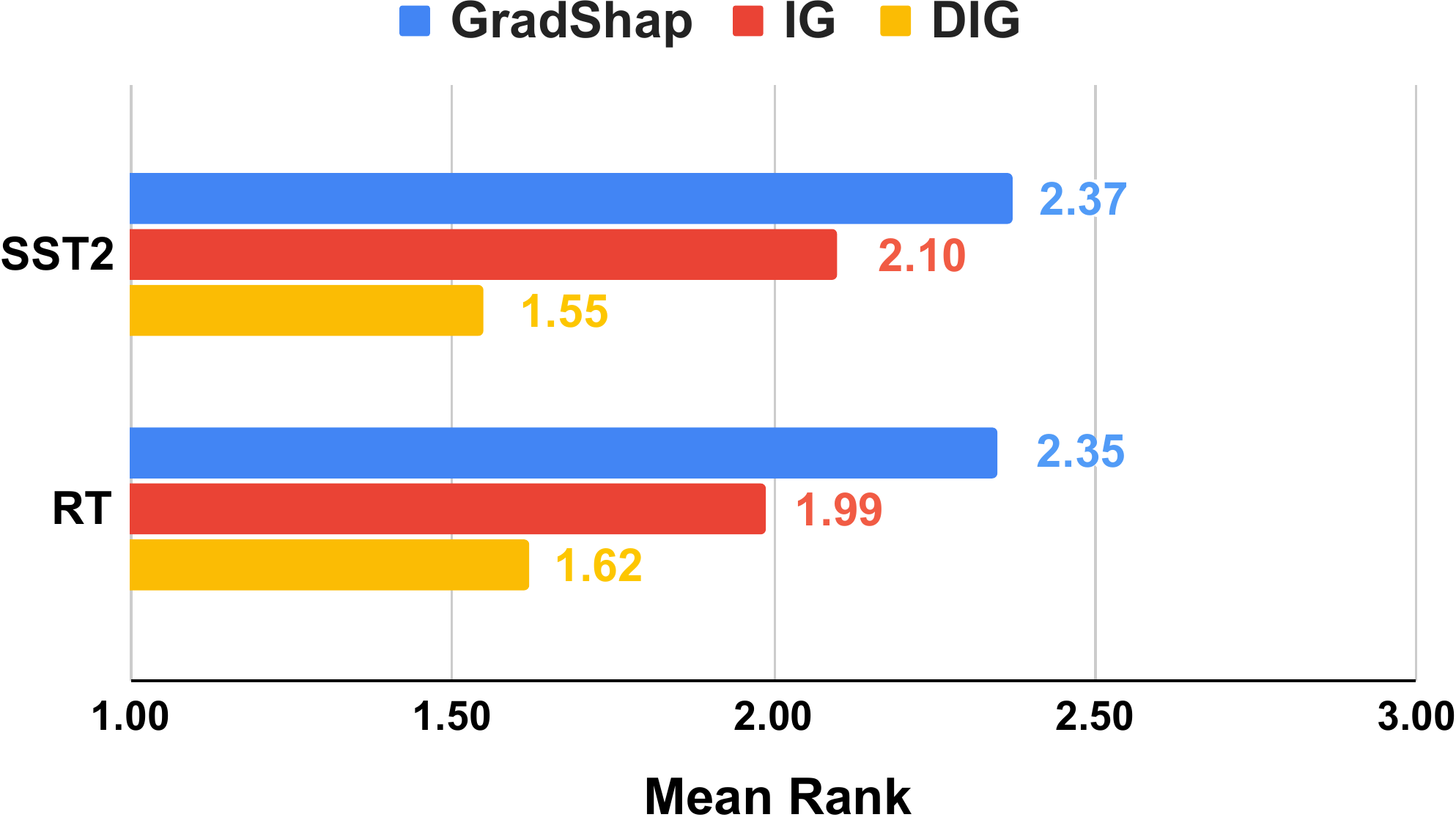}
	\vspace{-0.6cm}
	\caption{Result of human evaluation on DistilBERT model fine-tuned on SST2 dataset and BERT model fine-tuned on Rotten Tomatoes dataset. A lower mean rank means higher trustworthy explanation algorithm. For more details, refer to Section \ref{sec:human}}
	\label{fig:human_eval}
	\vspace{-0.2cm}
\end{figure}

\subsection{Human Evaluation}
\label{sec:human}
To further understand the impact of our algorithm on end users, we conduct human evaluations of explanations from our method and the two top baselines - IG and GradShap. We perform the study on the DistilBERT model fine-tuned on SST2 dataset and the BERT model fine-tuned on Rotten Tomatoes dataset. Further, we select the best variant of \method{} on each dataset for explanation comparisons. First, we pick 50 sample sentences from each dataset with lengths between 5 and 25 words for easier visualizations. Then, we convert the attributions from each method into word highlights, whose intensity is determined by the magnitude of the attributions. Finally, we show the highlighted sentence and the model's predicted label to the annotators and ask them to rank the explanations on a scale of 1-3, ``1'' being the most comprehensive explanation that best justifies the prediction. 

Figure \ref{fig:human_eval} shows the mean rank of each explanation algorithm across the two datasets. We find that \method{} has a significantly lower mean rank compared to IG ($p-$value less than 0.001 and $10^{-4}$ on SST2 and Rotten Tomatoes respectively). Thus, we conclude that explanations generated by \method{} are also trustworthy according to humans. Please refer to Appendix \ref{sec:visualizations} for visualizations and discussion on explanations generated by our methods.

\begin{table}[]
	\resizebox{\columnwidth}{!}{%
		\begin{tabular}{lccc}
			\toprule
			\multirow{1}{*}{\textbf{Method}} & \multicolumn{1}{c}{\textbf{SST2}} & \multicolumn{1}{c}{\textbf{IMDB}} & \multicolumn{1}{c}{\textbf{RT}}	\\
			\midrule
			IG						& -0.950				& -0.446				& -0.424				\\
			\midrule
			\small{\baselinea{}}			& -1.217 $\pm$ 0.024	& -0.834 $\pm$ 0.021 	& -0.474 $\pm$ 0.003	\\
			\small{\baselinen{}}			& -1.247 $\pm$ 0.013	& -0.854 $\pm$ 0.015 	& -0.460 $\pm$ 0.010	\\
			\midrule
			\method{} (best)		& \textbf{-1.259}		& \textbf{-0.878}		& \textbf{-0.501}		\\
			\bottomrule
		\end{tabular}%
	}
	\caption{\label{tab:ablation} Comparison of \method{} with two ablation variants - \baselinea{} and \baselinen{} on the DistilBERT model. We report 5-seed average log-odds score for the randomized methods. Please refer to Section \ref{sec:ablation} for more details.}
\end{table}

\subsection{Performance Analysis}
In this section, we report the ablation of \textsc{AnchorSearch} and the effect of path density on \method{}. Please refer to Appendix \ref{sec:ablation_extended} for ablations on neighborhood size and discussions on computational complexity.

\label{sec:ablation}
\paragraph{Ablation Study on \textsc{AnchorSearch}.}
We ablate our methods with two random variants - \baselinea{} and \baselinen{}, in which the \textsc{AnchorSearch} step uses a random anchor selection heuristic. Specifically, in \baselinea{}, the anchor is selected randomly from the complete vocabulary. Thus, this variant just ensures that the selected anchor is close to some word in the vocabulary which is not necessarily in the neighborhood. In contrast, the \baselinen{} selects the anchor randomly from the neighborhood without using our proposed heuristics \maxcount{} or \greedy{}. The log-odds metrics of IG, the two ablations, and our best variant of \method{} for DistilBERT fine-tuned individually on all three datasets are reported in Table \ref{tab:ablation}. We report 5-seed average for the randomized baselines. We observe that \baselinea{} improves upon IG on all three datasets. This shows that generating interpolation points close to the words in the vocabulary improve the explanation quality. Further, we observe that \baselinen{} improves upon \baselinea{} on log-odds metric. One reason could be that the words in a neighborhood are more semantically relevant to the original word, leading to more coherent perturbations for evaluating model gradients. Finally, we observe that, on average, our proposed method is better compared to selecting a random anchor in the neighborhood. This shows that our search strategies \maxcount{} and \greedy{} are indeed helpful.

\begin{figure}
    \centering
    \includegraphics[width=\columnwidth]{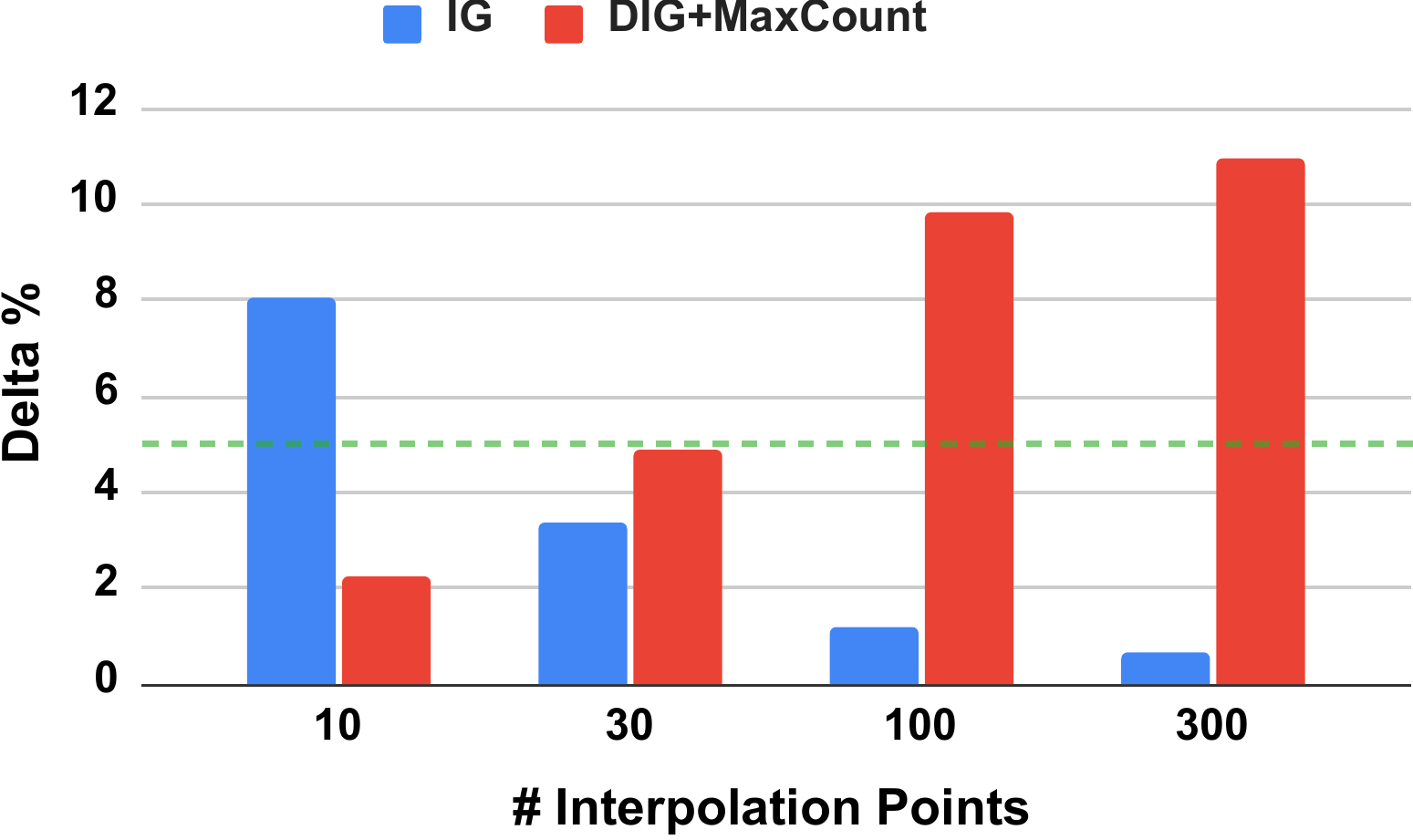}
    \caption{Effect of increasing number of interpolation points $m$ on IG and \method{}.}
    \label{fig:step_analysis}
\end{figure}

\begin{table}[t]
	\centering
	\resizebox{0.8\columnwidth}{!}{%
		\begin{tabular}{cccc}
			\toprule
			Factor $f$ & Log-Odds $\downarrow$ & WAE $\downarrow$ & Delta \% $\downarrow$	\\
			\midrule
			0		& -1.259 & 0.227	& 4.926	\\
			1		& -1.229 & 0.230	& 3.728	\\
			2		& -1.184 & 0.232	& 2.752	\\
			3		& -1.181 & 0.233	& 1.862	\\
			\bottomrule
		\end{tabular}%
	}
	\caption{\label{tab:density_analysis} Effect of up-sampling a path by a factor $f$ on Delta \% for \method{} using $m=30$.}
\end{table}

\paragraph{Effect of Increasing Path Density.}
In integrated gradients, the \textit{completeness} axiom (Section \ref{sec:axioms}) is used to estimate if the integral approximation (Equation \ref{eq:ig_sum}) error is low enough. This error is denoted as the Delta \% error. If the error is high, users can increase the number of interpolation points $m$.

While \method{} also satisfies the \textit{completeness} axiom, error reduction by increasing $m$ is infeasible. This is because increasing $m$ in Equation \ref{eq:dig_sum} implicitly changes the integral path rather than increasing the density. Hence, to achieve an error reduction in \method{}, we up-sample the interpolation path $P = \{w, w_1, w_2, \dots, w_{m-2}, w^{\prime}\}$ with an up-sampling factor ($f$) of one as follows:
\[
\resizebox{\columnwidth}{!}{
$P_1 = \{w, \frac{w + w_1}{2}, w_1, \frac{w_1 + w_2}{2}, \dots, \frac{w_{m-2} + w^{\prime}}{2}, w^{\prime}\},$
}
\]
i.e., we insert the mean of two consecutive points to the path. This essentially doubles the density of points in the path.
Similarly, $P_2$ can be obtained by up-sampling $P_1$, etc. \method{}($m,f=0$) refers to the standard \method{} with no up-sampling.


Given that we have two hyperparameters $m$ and $f$ that determine the overall path density, we analyze the effect of each of these in Figure \ref{fig:step_analysis} and Table \ref{tab:density_analysis} respectively. The results are shown for \method{}-\maxcount{} applied on DistilBERT model finetuned on SST2 dataset. In Figure \ref{fig:step_analysis}, we observe that as $m$ increases, the Delta \% of IG decreases as expected. But the trend is opposite for \method{}. As discussed above, for \method{}, the path length increases with increasing $m$, and hence, we attribute this trend to increasing difficulty in effectively approximating the integral for longer paths. Next, in Table \ref{tab:density_analysis}, we observe that as the up-sampling factor $f$ increases, the Delta \% consistently decreases. We also find that our up-sampling strategy does not increase the WAE by a significant amount with increasing $f$, which is desirable. Thus, this confirms that our up-sampling strategy is a good substitute of increasing $m$ for IG to effectively reduce the integral approximation error Delta \%. Following \citet{sundararajan2017axiomatic}, we choose a threshold of 5\% average Delta to select the hyperparameters. For more discussions, please refer to Appendix \ref{sec:density_details}.

\section{Related Works}
There has been an increasing effort in developing interpretability algorithms that can help understand a neural network model's behavior by explaining their predictions \cite{doshivelez2017rigorous,gilpin2019explaining}. Attributions are a post-hoc explanation class where input features are quantified by scalar scores indicating the magnitude of contribution of the features toward the predicted label. Explanation algorithms that generate attributions can be broadly classified into two categories - model-agnostic algorithms, like LIME \cite{lime}, Input occlusion \cite{li2016understanding}, Integrated gradients \footnote{Note that IG is strictly not a model-agnostic algorithm since it is defined for neural networks, but we still classify it as one since the scope of this work is limited to working on neural networks.}\cite{sundararajan2017axiomatic}, SHAP \cite{lundberg2017unified}, etc. and model-dependent algorithms, like LRP \cite{binder2016layer}, DeepLIFT \cite{shrikumar2017learning}, CD \cite{murdoch2018beyond}, ACD \cite{singh2018hierarchical}, SOC \cite{jin2020Towards}, etc. While the model-agnostic algorithms can be used as black-box explanation tools that can work for any neural network architecture, for the latter, one needs to understand the network's architectural details to implement the explanation algorithm. Typically, model-dependent algorithms require specific layer decomposition rules \cite{ancona2017towards,murdoch2018beyond} which needs to be defined for all the components in the model. Model-agnostic methods usually work directly with the model outputs and gradients which are universally available.

Due to the many desirable explanation axioms and ease of gradient computation, there has been several extensions of integrated gradients. For example, \citet{miglani2020investigating} study the effect of saturation in the saliency maps generated by integrated gradients. \citet{merrill2019generalized} extend integrated gradients to certain classes of discontinuous functions in financial domains. Further, \citet{jha2020enhanced} use KNNs and auto-encoders to learn latent paths for RNAs. Different from prior work, our focus here is to improve integrated gradients specifically for the discrete textual domain. While the idea of learning latent paths for text data is quite interesting, it brings a significant amount of challenge in successfully modeling such a complex latent space and hence, we leave this for future work.

\section{Conclusion}
In this paper, we proposed Discretized integrated gradients (\method{}) which is effective in explaining models working with discrete text data. Further, we proposed two interpolation strategies - \method{}-\greedy{} and \method{}-\maxcount{} that generate non-linear interpolation paths for word embedding space. Finally, we established the effectiveness of \method{} over integrated gradients and other gradient-based baselines through experiments on multiple language models and datasets. We also conduct human evaluations and find that \method{} enhances human trust on model predictions.

\bibliography{anthology,acl2020_updated}
\bibliographystyle{acl_natbib}

\clearpage

\appendix
\section{Preliminaries}
\label{sec:background}

\subsection{Attribution-based Explanations}
Attribution-based explanations generate a scalar score for a given input feature that indicates the contribution (or importance) of that feature towards particular label \cite{ancona2017towards}. Formally, let $x=\left[x_{1}, \ldots, x_{N}\right] \in \mathbb{R}^{N}$ be an input to a model which produces an output $y = [y_1, \ldots, y_C]$, where $C$ is the total number of labels. For a given label (usually the label predicted by the model), attribution-based explanation methods compute the contribution $R^{c}=\left[R_{1}^{c}, \ldots, R_{N}^{c}\right] \in \mathbb{R}^{N}$ of each feature.

\subsection{Integrated gradients}
\label{sec:background_ig}
Integrated gradients (IG) \cite{sundararajan2017axiomatic} for an input $x$ along the $i^{th}$ dimension is defined as follows:
\begin{equation}
\label{eq:ig_intergral}
\resizebox{\columnwidth}{!}{
$\textrm{ IG }_{i}(x) =\left(x_{i}-x_{i}^{\prime}\right) \times \int_{\alpha=0}^{1} \frac{\partial F\left(x^{\prime}+\alpha \times\left(x-x^{\prime}\right)\right)}{\partial x_{i}} d \alpha.$
}
\end{equation}
Here, $F$ is the neural network, $x^{\prime}$ is a baseline embedding, and $\alpha$ is the step size. Simply put, integrated gradients algorithm works by sampling points at a uniform spacing along a straight-line between the input and the baseline, and summing the model's gradient at the inputs for each interpolated points. To compute this integral efficiently, the authors propose a Riemann summation approximation defined below:
\begin{equation}
\small
\label{eq:ig_sum}
\resizebox{\columnwidth}{!}{
$\textrm{IG}^{\textrm{approx}}_{i}(x) = \left(x_{i}-x_{i}^{\prime}\right) \times \Sigma_{k=1}^{m} \frac{\left.\partial F\left(x^{\prime}+\frac{k}{m} \times\left(x-x^{\prime}\right)\right)\right)}{\partial x_{i}} \times \frac{1}{m},$
}
\end{equation}
where $m$ is the total number of steps considered for the approximation.

Next, we briefly describe how IG is used to explain a model's prediction which takes a sentence as input (for example, the model can be a text classification network).  Let $S = [w_0..w_n]$ be a sentence of length $n$ and $w_i$ be the $i^{th}$ word embedding of the sentence. Also, let $F$ be a text-classification model, i.e., $y = F(S)$. Then, IG calculates the attribution for each dimension of a word embedding $w_i$. The interpolation points required for Equation \ref{eq:ig_sum} are generated by linearly interpolating the word embedding between $w_i$ and a baseline word embedding (usually chosen as the pad embedding). Then, using Eq. \ref{eq:ig_sum}, the attribution for the $i^{th}$ dimension of $w$ is calculated. The final word attribution is the sum of the attributions for each dimension of the word embedding.

\section{Comparison with Integrated gradients and Path methods}
\label{sec:path}
It is easy to see that the approximation of integrated gradients is a special case of \method{}. Note that the $k^{th}$ linear interpolation of the $i^{th}$ dimension of input $x$ for IG can be represented as:
\begin{equation}
\label{eq:interpolate}
x^k_i = x^{\prime}_i + \frac{k}{m} \times (x_i - x^{\prime}_i).
\end{equation}
Substituting Eq. \ref{eq:interpolate} in Eq. \ref{eq:dig_sum} gives us Eq. \ref{eq:ig_sum}.

\citet{sundararajan2017axiomatic} define path methods as the general form of integrated gradients that are applicable for all monotonic paths between the input and the baseline. Our \method{} approach is a reformulation of the path method where the paths are not necessarily parameterized by $\alpha$, making it more applicable for discrete data domain. Hence, \method{} also satisfies all the theoretical properties applicable for path methods - Implementation Invariance, Sensitivity, Linearity, and Completeness. We refer the readers to Proposition 2 in \citet{sundararajan2017axiomatic} for more technical details.

\section{Evaluation Metrics}
\label{sec:eval_metrics_details}
In this section, we redefine the evaluation metrics and state the formulations for each of them. In this work, we use the following three automated metrics:
\begin{itemize}
	\item \textbf{Log-odds} (LO) score \cite{shrikumar2017learning} is defined as the average difference of the negative logarithmic probabilities on the predicted class before and after masking the top $k\%$ features with zero padding. Given the attribution scores generated by an explanation algorithm, we select the top k\% words based on their attributions replace them with zero padding. More concretely, for a dataset with $N$ sentences, it is defined as:
	$$
	\log -\operatorname{odds}(k)=\frac{1}{N} \sum_{i=1}^{N} \log \frac{p\left(\hat{y} \mid \boldsymbol{x}_{i}^{(k)}\right)}{p\left(\hat{y} \mid \boldsymbol{x}_{i}\right)},
	$$
	where $\hat{y}$ is the predicted class, $\boldsymbol{x}_{i}$ is the $i^{th}$ sentence, and $\boldsymbol{x}_{i}^{(k)}$ is the modified sentence with top k\% words replaced with zero padding. Lower scores are better.
	
	\item \textbf{Comprehensiveness} (Comp) score \cite{eraser} is the average difference of the change in predicted class probability before and after removing the top $k\%$ features. Similar to Log-odds, this measures the influence of the top-attributed words on the model's prediction. It is defined as:
	$$
	\operatorname{Comp}(k)=\frac{1}{N} \sum_{i=1}^{N} p(\hat{y} \mid \boldsymbol{x}_{i}^{(k)}) - p(\hat{y} \mid \boldsymbol{x}_{i}).
	$$
	Here $\boldsymbol{x}_{i}^{(k)}$ denotes the modified sentence with top $k\%$ words deleted from the sentence. Higher scores are better.

	\item \textbf{Sufficiency} (Suff) score \cite{eraser} is defined as the average difference of the change in predicted class probability before and after keeping only the top $k\%$ features. This measures the adequacy of the top $k\%$ attributions for model's prediction. It is defined in a similar fashion as comprehensiveness, except the $\boldsymbol{x}_{i}^{(k)}$ is defined as the sentence containing only the top $k\%$ words. Lower scores are better.
\end{itemize}

\begin{figure}
	\centering
	\begin{subfigure}{.5\textwidth}
		\centering
		\includegraphics[width=\columnwidth]{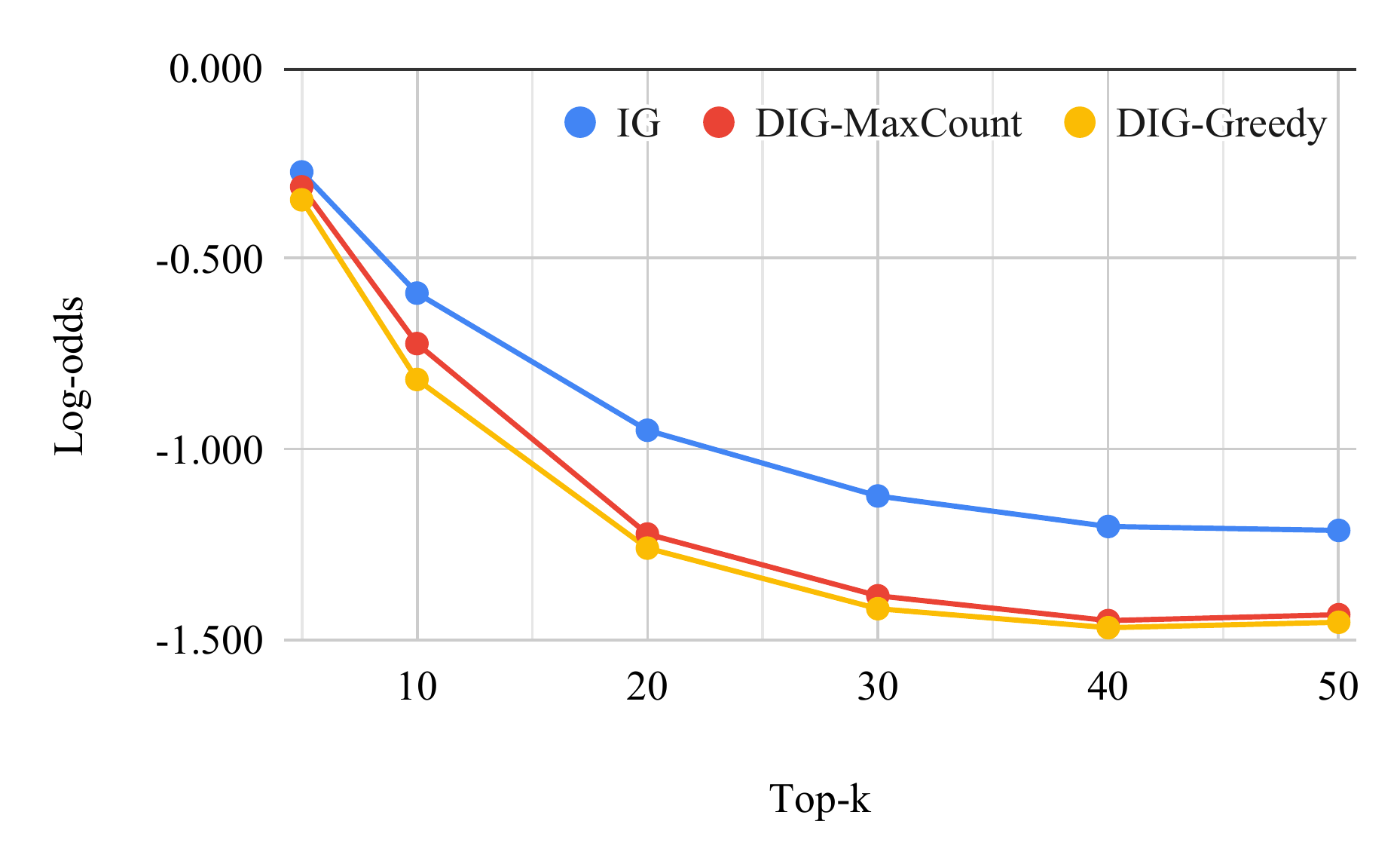}
		\caption{Log-odds $\downarrow$}
	\end{subfigure}%
\\
	\begin{subfigure}{.5\textwidth}
		\centering
		\includegraphics[width=\columnwidth]{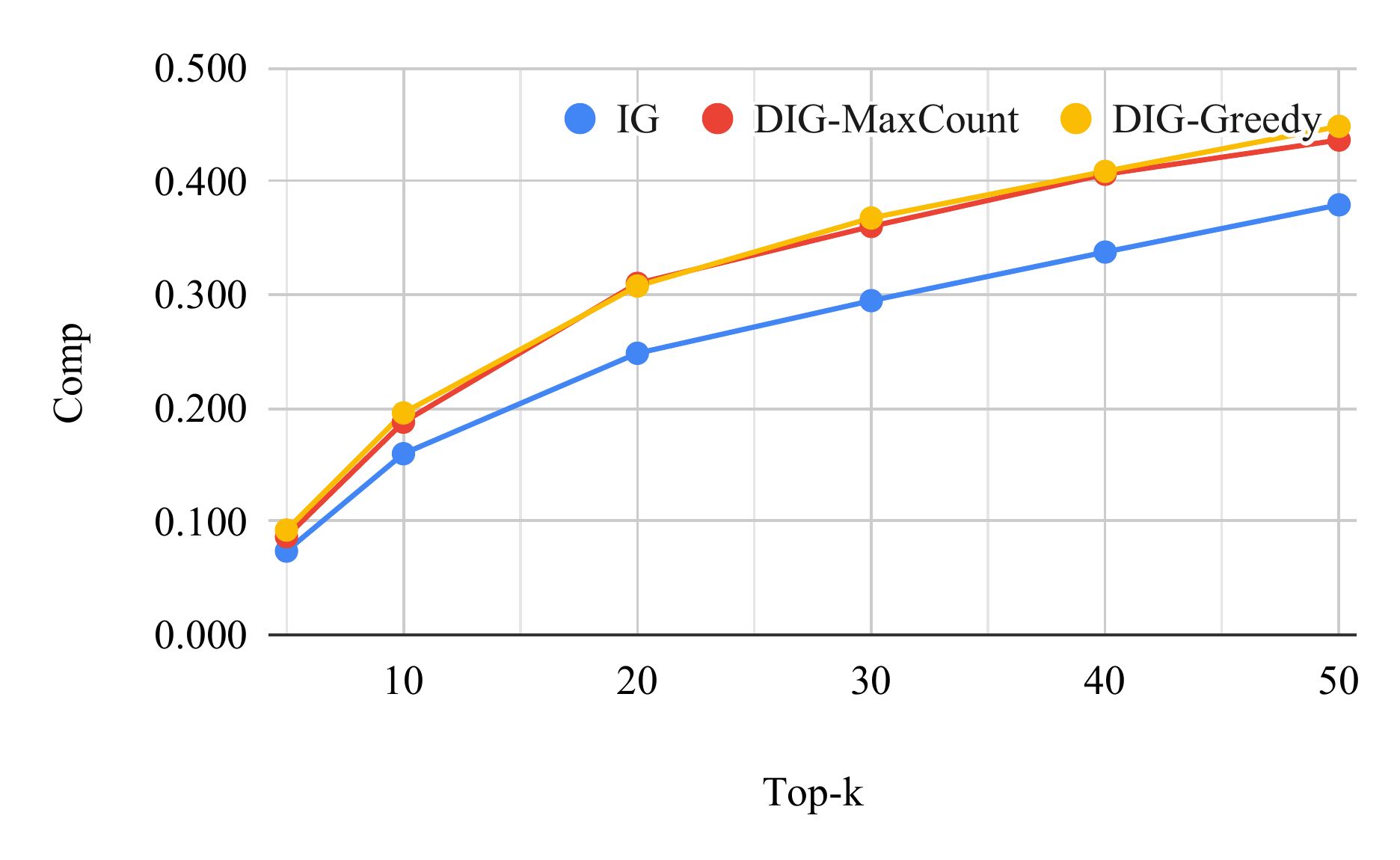}
		\caption{Comprehensiveness $\uparrow$}
	\end{subfigure}%
\\
	\begin{subfigure}{.5\textwidth}
		\centering
		\includegraphics[width=\columnwidth]{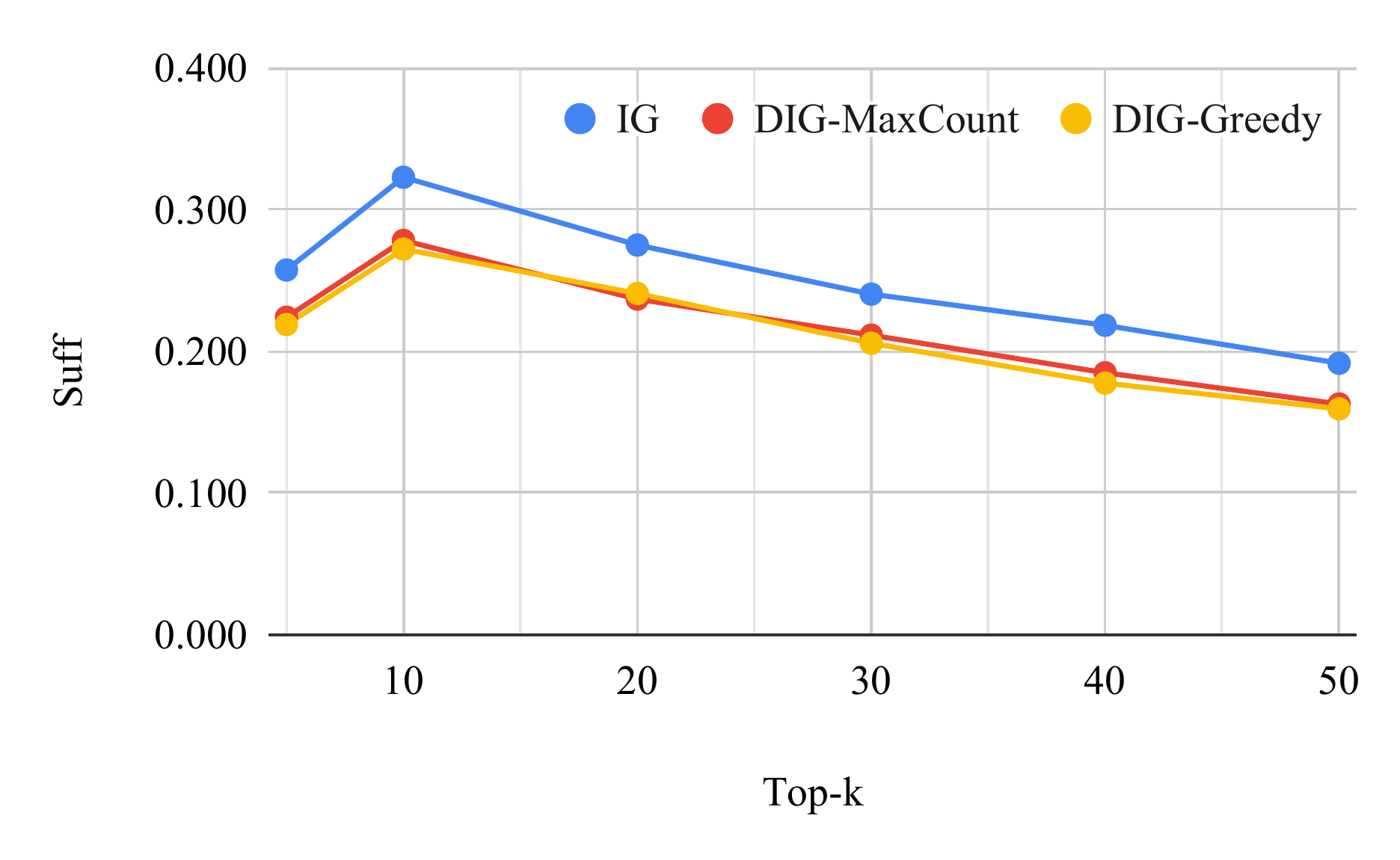}
		\caption{Sufficiency $\downarrow$}
	\end{subfigure}
	\caption{\label{fig:topk}Effect of changing top-k\% in log-odds, comprehensiveness, and sufficiency metric for the DistilBERT model fine-tuned on SST2 dataset.}
\end{figure}

\section{Effect of top-k in evaluation metrics}
\label{sec:topk}
In Figure \ref{fig:topk}, we visualize the effect of changing top-k\% on log-odds, comprehensiveness, and sufficiency metrics for DistilBERT model fine-tuned on the SST2 dataset. We compare the two variants of our method: \method{}-\greedy{} and \method{}-\maxcount{} with Integrated Gradients. We observe that our method outperforms IG for all values of k. Specifically, we note that the gap between \method{} and IG is initially non-existent but then gradually increases with increasing k in Figure \ref{fig:topk} (a) and eventually saturates. This shows that although IG might be equally good as \method{} at finding the top-5\% important words, the explanations from IG are significantly misaligned from true model behavior for higher top-k values.

\begin{table}[]
	\resizebox{\columnwidth}{!}{%
		\begin{tabular}{lccc}
			\toprule
			\multirow{1}{*}{\textbf{Dataset}} & \multicolumn{1}{c}{\textbf{DistilBERT}} & \multicolumn{1}{c}{\textbf{RoBERTa}} & \multicolumn{1}{c}{\textbf{BERT}}	\\
			\midrule
			SST2				& 1.00 & 0.00 & 0.42	\\
			IMDB				& 0.98 & -0.68 & 0.51	\\
			Rotten Tomatoes		& 0.21 & 0.22 & 0.23	\\
			\bottomrule
		\end{tabular}%
	}
	\caption{\label{tab:correlation} Pearson correlation between log-odds and WAE metrics for different dataset+LM settings. Please refer to Appendix \ref{sec:correlation} for more details.}
\end{table}

\section{Correlation between Log-odds and WAE}
\label{sec:correlation}
We compute the Pearson correlation between log-odds and WAE for each dataset + LM pair. For this, we consider the metric values for IG, \method{}-\greedy{}, and \method{}-\maxcount{} and report the correlations for each setting in Table \ref{tab:correlation}. We observe that, there is a strong correlation on average for DistilBERT. For BERT and RoBERTa we find a weak positive and negative correlation respectively.

\section{Ablation Studies}
\label{sec:ablation_extended}

\subsection{Effect of increasing path density}
\label{sec:density_details}
Here, we report the detailed analysis of the effect of increasing $m$ and $f$ in Tables \ref{tab:step_analysis} and \ref{tab:density_analysis_appendix} respectively. In Table \ref{tab:step_analysis}, we report the Log-odds score along with Delta \%. We do not note any consistent trend in Log-odds with increasing $m$ for both IG and \method{}. The results of IG suggest that, as long as the Delta \% is \textit{sufficiently} low, decreasing Delta \% any further doesn't impact the explanations very significantly. Further, in Table \ref{tab:density_analysis_appendix}, we report the WAE metrics to emphasize that our up-sampling strategy doesn't increase the WAE by a significant amount, which is desirable. Also, we note a consistent increase (although marginally) in Log-odds with decreasing Delta \%. But per our previous observations on IG, we believe these changes do not imply a causal relation between the two.

\begin{table}[]
	\centering
	\resizebox{\columnwidth}{!}{%
		\begin{tabular}{lcccc}
			\toprule
			\multirow{2}{*}{$m$}		& \multicolumn{2}{c}{IG} & \multicolumn{2}{c}{DIG} \\
			\cmidrule(r){2-3} \cmidrule(r){4-5}
			& Log-Odds $\downarrow$ & Delta \% $\downarrow$ & Log-Odds $\downarrow$ & Delta \% $\downarrow$	\\
			\midrule
			10		& -0.984 & 8.064 & -1.252 & 2.263	 \\
			30		& -0.950 & 3.394 & -1.259 & 4.926	 \\
			100		& -0.933 & 1.235 & -1.258 & 9.849	 \\
			300		& -0.940 & 0.703 & -1.242 & 10.955	 \\
			\bottomrule
		\end{tabular}%
	}
	\caption{\label{tab:step_analysis} Effect of increasing number of interpolation points $m$ on Delta \% for IG and \method{}. Please refer to Appendix \ref{sec:density_details} for more details.}
\end{table}

\begin{table}[]
	\centering
	\resizebox{\columnwidth}{!}{%
		\begin{tabular}{lccc}
			\toprule
			Up-sampling factor $f$ & Log-Odds $\downarrow$ & WAE $\downarrow$ & Delta \% $\downarrow$	\\
			\midrule
			\method{} ($m=30, f=0$)		& -1.259 & 0.227	& 4.926	\\
			\method{} ($m=30, f=1$)		& -1.229 & 0.230	& 3.728	\\
			\method{} ($m=30, f=2$)		& -1.184 & 0.232	& 2.752	\\
			\method{} ($m=30, f=3$)		& -1.181 & 0.233	& 1.862	\\
			\bottomrule
		\end{tabular}%
	}
	\caption{\label{tab:density_analysis_appendix} Effect of up-sampling a path by a factor $f$ on Delta \% for \method{}. For more details, refer to Appendix \ref{sec:density_details}.}
\end{table}

\begin{table}[]
	\centering
	\resizebox{0.75\columnwidth}{!}{%
		\begin{tabular}{lccc}
			\toprule
			\multirow{1}{*}{K} & \multicolumn{1}{c}{Log-odds $\downarrow$} & \multicolumn{1}{c}{WAE $\downarrow$} & \multicolumn{1}{c}{Delta \% $\downarrow$}	\\
			\midrule
			10		& -1.258 & 0.276 & 21.405	\\
			30		& -1.263 & 0.310 & 12.228	\\
			100		& -1.277 & 0.276 & 14.155	\\
			200		& -1.194 & 0.295 & 10.647	\\
			300		& -1.216 & 0.286 & 8.523	\\
			500		& -1.259 & 0.227 & 4.926	\\
			\bottomrule
		\end{tabular}%
	}
	\caption{\label{tab:nbrhood_analysis} Effect of increasing the neighborhood size $K$ of $KNN_V$ for \method{}. Please refer to Appendix \ref{sec:nbrhood_analysis} for more details.}
\end{table}

\subsection{Effect of increasing neighborhood size}
\label{sec:nbrhood_analysis}
In this section, we study the effect of increasing the neighborhood size in \method{}. The results are shown in Table \ref{tab:nbrhood_analysis}. We observe a clear decreasing trend in Delta \% with increasing neighborhood size, but there is no clear trend on Log-odds or WAE. Hence, we believe that the neighborhood size has little impact on the explanation quality, but we should still ensure sufficiently low Delta.

\subsection{Discussion on computational complexity}
In this section, we briefly discuss the computational complexity of our proposed interpolation strategies. The algorithms for \method{}-\greedy{} and \method{}-\maxcount{} are presented in Algorithms \ref{algo:dig_greedy} and \ref{algo:dig_count} respectively. From there, we observe that both our algorithms have a running time complexity of $\bigo(nmK)$, where $n$ is the number of words, $m$ is the number of interpolation points, and $K$ is the $KNN_V$ neighborhood size. While it is computationally feasible to parallelize the loops corresponding to $n$ and $K$, the same cannot be said for the loop corresponding to $m$ because we select the interpolation points iteratively. Although we empirically find in Section \ref{sec:density_details} that a small number of interpolation points are sufficient to calculate the explanations, we believe this bottleneck can be further tackled through efficient design of non-iterative search algorithms. We leave this for future works.

\section{Visualizations of explanations}
\label{sec:visualizations}
\begin{figure*}
	\centering
	\begin{subfigure}{\textwidth}
		\centering
		\includegraphics[width=\columnwidth]{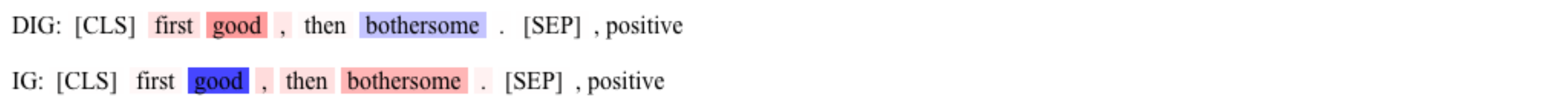}
	\end{subfigure}%
\vspace{1em}
\\
	\begin{subfigure}{\textwidth}
		\centering
		\includegraphics[width=\columnwidth]{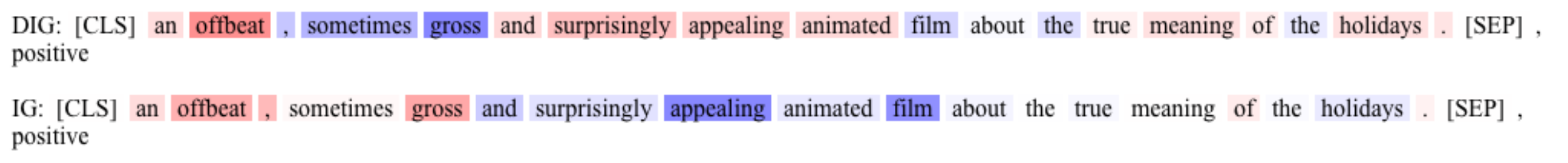}
	\end{subfigure}%
\vspace{1em}
\\
	\begin{subfigure}{\textwidth}
		\centering
		\includegraphics[width=\columnwidth]{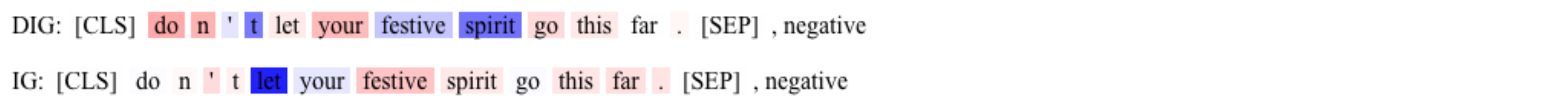}
	\end{subfigure}
\vspace{1em}
\\
	\begin{subfigure}{\textwidth}
		\centering
		\includegraphics[width=\columnwidth]{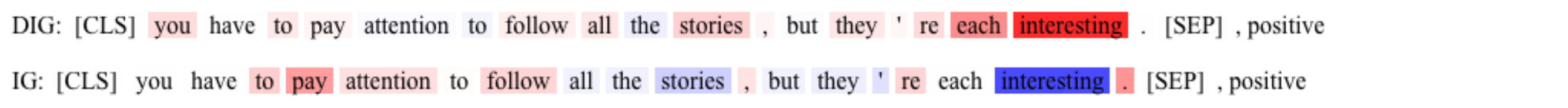}
	\end{subfigure}
\vspace{1em}
\\
	\begin{subfigure}{\textwidth}
		\centering
		\includegraphics[width=\columnwidth]{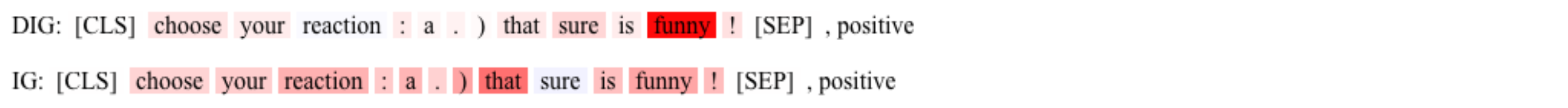}
	\end{subfigure}
\vspace{1em}
\\
	\begin{subfigure}{\textwidth}
		\centering
		\includegraphics[width=\columnwidth]{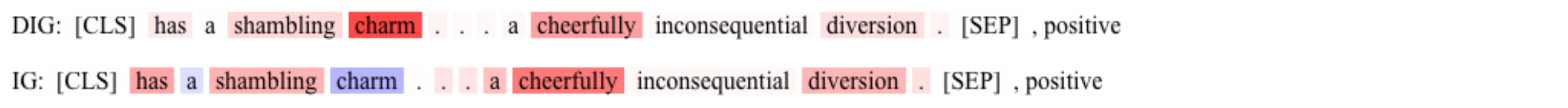}
	\end{subfigure}
\vspace{1em}
\\
	\begin{subfigure}{\textwidth}
		\centering
		\includegraphics[width=\columnwidth]{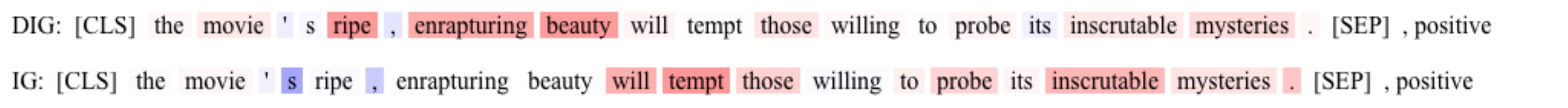}
	\end{subfigure}
\vspace{1em}
\\
	\begin{subfigure}{\textwidth}
		\centering
		\includegraphics[width=\columnwidth]{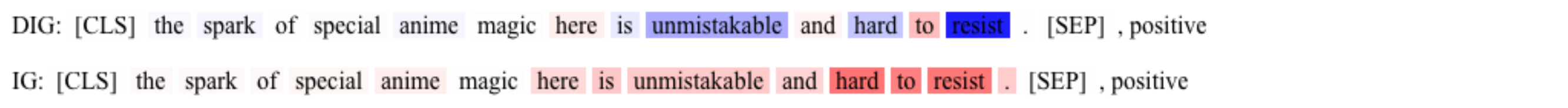}
	\end{subfigure}
\vspace{1em}
\\
	\begin{subfigure}{\textwidth}
		\centering
		\includegraphics[width=\columnwidth]{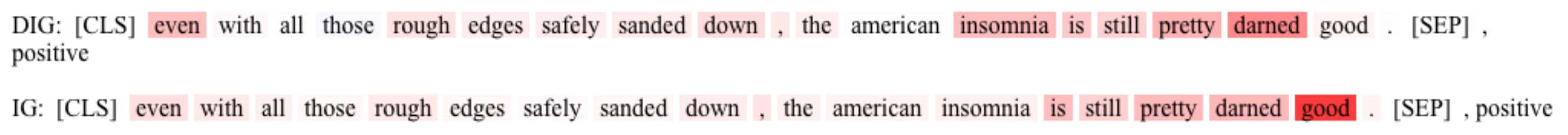}
	\end{subfigure}
\vspace{1em}
\\
	\begin{subfigure}{\textwidth}
		\centering
		\includegraphics[width=\columnwidth]{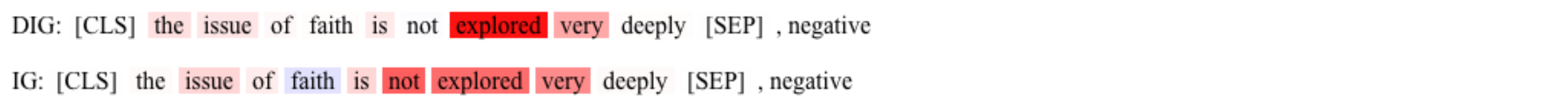}
	\end{subfigure}
\vspace{1em}
	\caption{\label{fig:example_visuals} Some example visualizations of attributions from \method{} and IG for the DistilBERT model fine-tuned on SST2 dataset. The sentence visualization is followed by model's sentiment prediction for the sentence. Here, the red highlighted words denote positive attributions and blue denotes negative attributions. For more details, please refer to Appendix \ref{sec:visualizations}}
\end{figure*}

In this section, we present some interesting sentence visualizations based on explanations from \method{} and IG for SST2 dataset in Figure \ref{fig:example_visuals}. We show the sentence visualization and the model's predicted sentiment for the sentence for each explanation algorithm. In the visualizations, the red highlighted words denote positive attributions and blue denotes negative attributions. That is, the explanation model suggests that the red highlighted words support the predicted label whereas the blue ones oppose (or undermine) the prediction. We observe that in many cases, \method{} is able to highlight more plausible explanations. For example, in sentence pairs 1-7, clearly the \method{} highlights are more inline with the model prediction. But we want to emphasize that it does not mean that our method always produces more plausible highlights. For example, for sentences 8-10, we observe that highlights from IG are more plausible than those of \method{}. Hence, this shows that, while it could be a good exercise to visualize the attributions as a sanity check, we should rely more on automated metrics and human evaluations to correctly compare explanation algorithms.

\begin{algorithm}
	\SetKwFunction{monotonize}{\textsc{Monotonize}}
	\SetKwFunction{distance}{Distance}
	\SetKwInOut{KwIn}{Input}
	\SetKwInOut{KwOut}{Output}
	
	\KwIn{Sentence $s = [w_1, w_2, \dots w_n]$, $k$-nearest neighbor graph for the vocabulary $KNN_V$, number of interpolation points $m$}
	\KwOut{Interpolations}
	
	$points = [\ ]_{n*m}$
	
	\For{$i \leftarrow 1$ \KwTo $n$}{
		\For{$j \leftarrow 1$ \KwTo $m$}{
		    $dists = \{\ \}$\\
		    \For{$k \leftarrow 1$ \KwTo $K$}{
		        $nbr \leftarrow KNN_V(w_i)[k]$\\
		        $c' \leftarrow \monotonize(nbr, w_i)$\\
		        $dists[nbr] \leftarrow \distance(nbr, c')$\\
			}
			$a \leftarrow \argmin_{a' \in dists} {dists[a']}$\\
			$c \leftarrow \monotonize(a, w_i)$\\
			$points[i, j] \leftarrow c$
		}
	}
	
	\KwRet{$points$}
	\caption{\label{algo:dig_greedy}\method{}-\greedy{}}
\end{algorithm}

\begin{algorithm}
	\SetKwFunction{monotonize}{\textsc{Monotonize}}
	\SetKwInOut{KwIn}{Input}
	\SetKwInOut{KwOut}{Output}
	
	\KwIn{Sentence $s = [w_1, w_2, \dots w_n]$, $k$-nearest neighbor graph for the vocabulary $KNN_V$, number of interpolation points $m$}
	\KwOut{Interpolations}
	
	$points = [\ ]_{n*m}$
	
	\For{$i \leftarrow 1$ \KwTo $n$}{
		\For{$j \leftarrow 1$ \KwTo $m$}{
			$a \leftarrow \argmax_{a' \in KNN_V(w_i)} {|M_{a'}|}$\\
			$c \leftarrow \monotonize(a, w_i)$\\
			$points[i, j] \leftarrow c$
		}
	}
	
	\KwRet{$points$}
	\caption{\label{algo:dig_count} \method{}-\maxcount{}}
\end{algorithm}

\end{document}